# Application of Grey Numbers to Assessment Processes

[1] Michael Gr. Voskoglou, [2] Yiannis A. Theodorou

[1] Professor Emeritus of Mathematical Sciences
School of Technological Applications
Graduate Technological Educational Institute of Western Greece, Patras, Greece
E-mail: mvosk@hol.gr , voskoglou@teiwest.gr

[2] Professor of Mathematics,
Department of Electronic Engineering
Graduate Technological Educational Institute of Central Greece, Lamia, Greece
E-mail: teo@teiste.gr

## Abstract

*The theory of grey systems plays an important role in science, engineering and in the everyday life in general for handling approximate data. In the present paper grey numbers are used as a tool for assessing with linguistic expressions the mean performance of a group of objects participating in a certain activity. Two applications to student and football player assessment are also presented illustrating our results.*

## 1. Introduction

The assessment of a system's effectiveness (i.e. of the degree of attainment of its targets) with respect to an action performed within the system (e.g. problem-solving, decision making, learning, etc) is a very important task, which enables the correction of the system's weaknesses resulting to the improvement of its general performance. The assessment methods that are usually applied in practice are based on principles of the bivalent logic (yes-no). However, situations appear frequently in real life characterized by a degree of uncertainty and/or ambiguity. For example, a teacher is frequently not sure for a particular numerical grade characterizing a student's performance. In such cases fuzzy logic, due to its property of characterizing the ambiguous/uncertain situations with multiple values, offers rich resources for the assessment purposes.

The traditional method for assessing a group's *mean performance* is the calculation of the *mean value* of its members' numerical scores. However, frequently in practice the individual performance is evaluated not by numerical scores, but by qualitative characterizations (grades), like excellent, very good, good, fair, unsatisfactory, etc. In such cases the calculation of the mean values can not be applied, but one could use the also traditional method of calculating the *Grade Point Average (GPA) index* (e.g. see [4], p. 125) for assessing the group's performance. However GPA actually measures not the mean, but the group's *quality performance*, by assigning greater coefficients to the higher scores.

In earlier works we have used the measurement of a fuzzy system's *uncertainty* as a tool for assessing a group's mean performance (e.g. see [4],: Chapter 5). Nevertheless, this method, apart of requiring laborious calculations, can be used to compare the mean performance of two different groups only under the assumption that the initially existing uncertainty for the two groups is the same, a condition which is not always true.

More recently we have also used *fuzzy numbers* as tools for evaluating a group's mean performance (e.g. see [4],: Chapter 7, [5], etc.) that appears to be a more general and accurate method than the measurement of the uncertainty.

In the paper at hand we develop an analogous assessment method by using *grey numbers (GNs)* instead of fuzzy numbers. The rest of the paper is formulated as follows: In Section 2 we introduce the necessary for our purposes background from the theory of GNs. In Section 3 we develop the new assessment method, while in Section 4 we present two numerical examples on assessing student and player performance. The paper closes with our conclusion and hints for further research.

## 2. Grey Numbers

Frequently in the everyday life, as well as in many applications of science and engineering including medicine diagnostics, psychology, sociology, control systems, economy price indices, opinion polls, etc., the data can not be easily determined precisely and in practice estimates of them are used. Apart from fuzzy logic, another effective tool for handling the approximate data is the use of the GNs, which are introduced with the help of the real intervals.

A GN is an indeterminate number whose probable range is known, but which has unknown position within its boundaries. Therefore, if $R$: denotes the set of real numbers, a GN, say A, can be expressed mathematically by

$$A \in [a, b] = \{x \in R: a \leq x \leq b\}.$$

Compared with the interval $[a, b]$ the GN A enriches its uncertainty representation with the *whitenization function*, defining a *degree of greyness* for each $x$ in $[a, b]$. When $a = b$, then A is called a *white number* and if $A \in (-\infty, +\infty)$, then A is called a *black number*. For general facts on GNs we refer to the book [2].

From the definition of the GNs it becomes evident that the well known arithmetic of the real intervals [3] can be used to define the basic arithmetic operations among the GNs. More explicitly, if $A_1 \in [a_1, b_1]$ and $A_2 \in [a_2, b_2]$ are given GNs, then we define:

- *Addition* by: $A_1 + A_2 \in [a_1 + a_2, b_1 + b_2]$
- *Subtraction* by: $A_1 - A_2 \in [a_1 - b_2, b_1 - a_2]$
- *Multiplication* by: $A_1 \times A_2 \in [\min\{a_1a_2, a_1b_2, b_1a_2, b_1b_2\}, \max\{a_1a_2, a_1b_2, b_1a_2, b_1b_2\}]$

- ***Division*** by: $A_1 : A_2 \in [\min\{\frac{a_1}{a_2}, \frac{a_1}{b_2}, \frac{b_1}{a_2}, \frac{b_1}{b_2}\}, \max\{\frac{a_1}{a_2}, \frac{a_1}{b_2}, \frac{b_1}{a_2}, \frac{b_1}{b_2}\}]$, where $0 \notin [a_2, b_2]$.

- ***Scalar multiplication*** by: $kA_1 \in [ka_1, kb_1]$, where k is a positive real number.

Let us denote by w(A) the white number with the highest probability to be the representative real value of the GN $A \in [a, b]$. The technique of determining the value of w(A) is called ***whitenization of A***.

One usually defines $w(A) = (1-t)a + tb$, with t in [0, 1]. This is known as ***equal weight whitenization***. When the distribution of A is unknown, we take $t = \frac{1}{2}$, which gives that $w(A) = \frac{a+b}{2}$.

## 3. The Assessment Method

Let G be a group of *n* objects participating in a certain activity. Assume that one wants to assess the mean performance of G in terms of the fuzzy linguistic expressions (grades) A = Excellent, B = Very good, C = Good, D = Fair and F = Unsatisfactory (Failed).

We introduce a numerical scale of scores from 0 to 100 and we correspond these scores to the linguistic grades as follows: A (100-85), B(84-75), C (74-60), D(59-50) and F(49-0). This correspondence, although it satisfies the common sense, it is not unique, depending on the user's personal goals. For example, in a more strict assessment one could take A (100-90), B(89-80), C (79-70), D(69-60) and F(59-0), etc.

It is possible now to represent each linguistic grade by a GN, denoted for simplicity with the same letter. Namely, we introduce the GNs: $A \in [85, 100]$, $B \in [75, 84]$, $C \in [60, 74]$, $D \in [50, 59]$ and $F \in [0, 49]$.

Let $n_A$, $n_B$, $n_C$, $n_D$ and $n_F$ denote the numbers of the objects of G whose performance was characterized by the grades A, B, C, D and F respectively.

Application of Grey Numbers to…..Assigning to each object of G the corresponding GN we define the *mean value* of all those GNs to be the GN:

$$M = \frac{1}{n}[n_A A + n_B B + n_C C + n_D D + n_F F].$$

It is logical now to consider the GN M as the representative of the group's mean performance. Since the distributions of the GNs A, B, C, D and F are unknown, the same happens with the distribution of M. Therefore, if $M \in [m_1, m_2]$, we can take $w(M) = \frac{m_1 + m_2}{2}$. It becomes evident that the value of w(M) determines the linguistic grade characterizing the group's mean performance.

*Remark:* In earlier works(e.g. see [5]) he have assessed the mean performance of G by using the *Triangular Fuzzy Numbers (TFNs)* A = (85, 92.5, 100}, B = (75, 79.5, 84), C = (60, 67, 74], D = (50, 54.5, 59] and F = (0, 24.5 49] instead of the corresponding GNs used here. Assigning to each object of G the TFN corresponding to its performance we calculated the mean value M (*a, b, c*) of all those TFNs and we proved that the representative real value defuzzifying M is equal to $b = \frac{a+c}{2}$. Consequently the two assessment methods (TFNs and GNs) are *equivalent* to each other.

## 4. Examples

The two real life applications presented here illustrate the assessment method developed in the previous section.

*Example 1:* The following Table depicts the performance of two student groups, say $G_1$ and $G_2$, in a common mathematical examination. Compare the mean performance of the two groups.

277

**Table 1:** Student performance

| Grade | $G_1$ | $G_2$ |
|---|---|---|
| A | 20 | 20 |
| B | 15 | 30 |
| C | 7 | 15 |
| D | 10 | 15 |
| F | 8 | 5 |
| Total | 60 | 85 |

Assigning to each student the corresponding GN we calculate the mean values $M_1$ and $M_2$ of all those GNs for the groups $G_1$ and G2 respectively, which are approximately equal to:

$$M_1 = \frac{1}{60}(20A+15B+7C+10D+8F) \in [62.42, 79.33]$$

$$M_2 = \frac{1}{85}(20A+30B+15C+15D+5F) \in [65.88, 79.53]$$

Therefore $w(M_1) \approx \frac{62.42+79.33}{2} \approx 70.88$ and $w(M_2) \approx \frac{65.88+79.53}{2} \approx 72.71$.

Consequently both groups demonstrated a good (C) mean performance, with the performance of the second group being better.

*Example 2:* The performance in a game of five players of a soccer club was assessed by six different athletic journalists using a scale from 0 to 100 as follows:

**$P_1$** (player 1)**:** 43, 48, 49, 49, 50, 52

**$P_2$:** 81, 83. 85, 88, 91, 95

**$P_3$:** 76, 82, 89, 95, 95, 98

**$P_4$:** 86, 86, 87, 87, 87, 88

**$P_5$:** 35, 40, 44, 52, 59, 62.

Find the mean performance of the five players using GNs.

Inspecting the given data one observes that the 30 in total scores assigned to the five players by the six journalists correspond to 14 characterizations of excellent (A) performance, to 4 for very good (B), to 1 for good (C), to 4 for fair (D) and to 7 characterizations for unsatisfactory (F) performance. Therefore, the mean player performance can be assessed by the GN

$$M = \frac{1}{30}(14A + 4B + C + 4D + 7F) \in [58.33, 79.63].$$

Therefore $w(M) = \frac{58.33 + 79.63}{2} = 68.98$, which shows that the five players demonstrated a good (C) mean performance.

***Remark:*** The mean value of the 30 in total scores assigned to the five players by the journalists is 72.06 demonstrating a good (C) mean performance. The difference of 72.06-68.98=3.06 units between the two assessment methods is due to the approximate character of the method with the GNs. In fact, consider the two extreme cases, where the maximal or the minimal possible numerical score is assigned to each player for each linguistic grade, e.g. all (the 14) scores corresponding to A, are 100 or 85 respectively, etc. Then, the mean values of the player scores are 79.63 and 58.33 respectively and w(M) is equal to the mean value of these two values. Consequently, the method with the GNs is useful only when no numerical scores are given and the group's performance is assessed by qualitative grades (as in Example 1).

## 5. Conclusion

In the paper at hand GNs were used as a tool for developing an assessment method of a group's performance during an activity. Although this method was proved to be equivalent with a corresponding method using TFNs developed in earlier works, the required computational burden was significantly reduced.

The theory of grey systems plays in general an important role in science, engineering and in the everyday life for handling approximate data. In [1], for example, the eigenvalue problem of Correspondence Analysis with grey data was studied. The development of further applications of grey systems to real life problems is one of the main components of our future research.

## References

**[1] Alevizos, P. D., Theodorou, Y. & Vrahatis, M.N. (2017)**, Correspondence Analysis with Grey Data: The Grey Eigen Value Problem, *The Journal of Grey System*, 29(1), 92-105.

**[2] Liu, S. F. & Lin, Y. (Eds.) (2010)**, *Advances in Grey System Research*, Springer, Berlin – Heidelberg.

**[3] Moore, R.A. , Kearfort, R. B. & Clood, M.J. (1995)**, *Introduction to Interval Analysis*, 2$^{nd}$ Printing, SIAM, Philadelphia.

**[4] Voskoglou, M. Gr. (2017),** *Finite Markov Chain and Fuzzy Logic Assessment Models; Emerging Research and Opportunities*, Createspace.com. – Amazon, Columbia, SC, USA.

**[5] Voskoglou, M. Gr. & Subbotin, I.Ya. (2017),** Application of Triangular Fuzzy Numbers for Assessing the Results of Iterative Learning, *International Journal of Applications of Fuzzy Sets and Artificial Intelligence*, 7, 59-72.